\documentclass[preprint,12pt]{elsarticle}

\usepackage{amssymb}
\usepackage[a4paper, total={6in, 8in}]{geometry}\usepackage{float}

\usepackage{booktabs}
  

\journal{arXiv}

\begin{document}

\begin{frontmatter}



\title{Surrogate Modeling of Melt Pool Thermal Field using Deep Learning}


\author[inst1]{AmirPouya Hemmasian}
\author[inst1]{Francis Ogoke}
\author[inst1]{Parand Akbari}
\author[inst1,inst2]{Jonathan Malen}
\author[inst1,inst2]{Jack Beuth}
\author[inst1,inst3,inst4]{Amir Barati Farimani}

\affiliation[inst1]{organization={Department of Mechanical Engineering, Carnegie Mellon University},
            city={Pittsburgh},
            postcode={15213}, 
            state={PA},
            country={USA}}
\affiliation[inst2]{organization={Department of Materials Science and Engineering, Carnegie Mellon University},
            city={Pittsburgh},
            postcode={15213}, 
            state={PA},
            country={USA}}
\affiliation[inst3]{organization={Department of Chemical Engineering, Carnegie Mellon University},
            city={Pittsburgh},
            postcode={15213}, 
            state={PA},
            country={USA}}
\affiliation[inst4]{organization={Machine Learning Department, Carnegie Mellon University},
            city={Pittsburgh},
            postcode={15213}, 
            state={PA},
            country={USA}}

\begin{abstract}
Powder-based additive manufacturing has transformed the manufacturing industry over the last decade. In Laser Powder Bed Fusion, a specific part is built in an iterative manner in which two-dimensional cross-sections are formed on top of each other by melting and fusing the proper areas of the powder bed. In this process, the behavior of the melt pool and its thermal field has a very important role in predicting the quality of the manufactured part and its possible defects. However, the simulation of such a complex phenomenon is usually very time-consuming and requires huge computational resources. Flow-3D is one of the software packages capable of executing such simulations using iterative numerical solvers. In this work, we create three datasets of single-trail processes using Flow-3D and use them to train a convolutional neural network capable of predicting the behavior of the three-dimensional thermal field of the melt pool solely by taking three parameters as input: laser power, laser velocity, and time step. The CNN achieves a relative Root Mean Squared Error of 2\% to 3\% for the temperature field and an average Intersection over Union score of 80\% to 90\% in predicting the melt pool area. Moreover, since time is included as one of the inputs of the model, the thermal field can be instantly obtained for any arbitrary time step without the need to iterate and compute all the steps.

\end{abstract}




\end{frontmatter}


\section{Introduction}

Metal additive manufacturing (MAM) has attracted increasing scientific attention from industry and academia owing to its noticeable advantages relative to conventional manufacturing methods \cite{debroy2018additive}. It enables the manufacturing of complex geometries with higher reliability and reduced material consumption and lead time. \cite{jiang2020path, AKBARI2022102817} Despite these benefits, controlling the quality of the printed parts is a major challenge as defects such as lack of fusion, keyhole, and balling can compromise the structural integrity of the additively manufactured components. \cite{WANG2020101538} Since these defects are rooted in the dynamics of the melt pool-- a locally melted zone generated by the emission of the laser on the powder particles-- monitoring and controlling the melt pool's geometry and thermal field are of critical importance. \cite{AKBARI2022102817} 

To monitor the melt pool in the Laser Powder Bed Fusion (LPBF) process, experiments can be conducted. However, they are expensive and need laborious preparation and adjustment. In addition, employing simulation methods such as Flow-3D Software\cite{FLOW-3D} to generate computational predictions in such a complex multi-scale and multi-physics phenomenon is time-consuming, and also requires huge computational resources and integration of multiple discrete and costly simulations. \cite{yan2018data} Flow-3D is one of the software packages capable of executing such simulations utilizing iterative numerical solvers.

As an alternative to the computationally expensive numerical solvers like Flow-3D, we leverage the power of deep learning to make fast and accurate predictions of the three-dimensional thermal field of the melt pool over time. In the last decade, deep learning has caused a revolution in the field of computer vision \cite{krizhevsky2012imagenet, mnih2015human}. Deep neural networks have shown success in many challenging areas like computer vision \cite{krizhevsky2012imagenet, voulodimos2018deep} and speech recognition \cite{hinton2012deep, deng2013new, graves2013speech, abdel2014convolutional}. Following the introduction of convolutional neural networks (CNNs) by Lecun et al.\cite{lecun1995convolutional} and the demonstrated capability of AlexNet on the image classification benchmark \cite{krizhevsky2012imagenet}, these networks have become widely used in many computer vision tasks including image recognition and segmentation\cite{simonyan2014very,long2015fully}, as well as video classification\cite{karpathy2014large}. Inspired by this, many researchers have been applying deep learning models in science and engineering problems \cite{kutz2017deep, kirchdoerfer2016data, oishi2017computational}. Deep learning has also been utilized to study complex physical systems by obtaining accurate solutions to their systems of governing equations, which are usually ordinary or partial differential equations\cite{sirignano2018dgm, han2018solving, raissi2019physics}. Due to the high computational cost of conventional numerical solvers, developing surrogate models to study physical systems has been of great interest to many researchers and scientists as well\cite{liang2018deep, jin2020deep,tang2020deep}. Convolutional neural networks are well-suited for dealing with spatially structured data due to their inherent translational invariance, weight sharing, and feature extraction properties. They have been applied to obtain rapid prediction and simulation of complex phenomena like reaction-diffusion systems \cite{li2020reaction} and convective fluid flows \cite{jiang2020deep}.

In this work, we approximate the thermal field and geometry of the melt pool during the LPBF process by utilizing the capabilities of CNNs. Our model's input is simply a vector containing three numbers: The laser beam's power and velocity, and the time step. It succeeds in learning the dominant pattern of the melt pool's thermal field for a wide variety of process parameters. We evaluate the performance of our model qualitatively by visual comparison with the results from Flow-3D, and quantitatively by measuring the Root Mean Squared Error for the thermal field and Intersection over Union for the melt pool region.

\section{Dataset} 

\renewcommand{\arraystretch}{1.5}
\begin{table}[htbp]
\caption{The datasets used in this work.}
\begin{center}
\begin{tabular}{|c|c|c|c|c|c|c|c|}
\hline
\textbf{name} & \textbf{alloy} & \textbf{mesh} & \textbf{powder} & \textbf{train} & \textbf{val} & \textbf{P(w)} & \textbf{V(mm/s)}\\
\hline
Ti64-5m & Ti-6Al-4V & $5 \mu m$ & No & 39 & 7 & 100-520 & 400-1200\\
\hline
Ti64-10m & Ti-6Al-4V & $10 \mu m$ & No & 287 & 71 & 100-460 & 100-1500\\
\hline
Ti64-10m-p & Ti-6Al-4V & $10 \mu m$ & Yes & 555 & 138 & 100-560 & 100-1400\\
\hline
SS-10m-p & SS316L & $5 \mu m$ & Yes & 32 & 7 & 100-320 & 100-400\\
\hline
\end{tabular}
\label{table:data}
\end{center}
\end{table}

A summary of the datasets created for this work can be found in Table \ref{table:data}. Each simulation contains 100 timesteps spaced at 5 $\mu s$ intervals. The FLOW-3D v11.2 CFD package was used to simulate the melting process \cite{FLOW-3D}. Generally, coupled PDEs describing the mass, momentum, and energy conservation present in the system are solved using a volume of fluid (VOF) model. These  equations are reproduced below:

\begin{equation}
    \nabla \cdot (\rho \vec{v}) = 0
\end{equation}

\begin{equation}
    \frac{\partial \vec{v} }{\partial t} + (\vec{v} \cdot \nabla) \vec{v} = - \frac{1}{\rho}\nabla \vec{P} + \mu \nabla^2 \vec{v} + \vec{g}(1-\alpha (T-T_m))
\end{equation}

\begin{equation}
\frac{\partial h}{\partial t} + \left ( \vec{v} \cdot \nabla \right ) h = \frac{1}{\rho}\left( \nabla \cdot k \nabla T \right) 
\end{equation}
where $\vec{P}$ is the pressure of the system, $\vec{g}$ represents gravity, $\alpha$ is the coefficient of thermal expansion, $\rho$ is the density of the material, $h$ is the specific enthalpy, $\vec{v}$ is the velocity of the system and $k$ is the heat conductivity of the material. The thermal parameters, $C_p$, $k$, as well as the density, $\rho$, are assumed to vary with temperature, with their functional relationships described in \cite{mills2002recommended} for both materials.  While the material is assumed to be incompressible in these simulations, however, there are accommodations for the variation of density with temperature. To do so, the incompressible flow algorithm is solved using a  variable density, calculated based on the local temperature values.

To compute the dynamics of the free surface of the melt pool, the VOF method is used to define the fluid configuration \cite{hirt1981volume}. A conservation equation can be written for the Volume of Fluid function, $F(x,y,z,t)$, as demonstrated in equation \ref{VOF}. When $F = 1$, the mesh cell contains fluid, whereas when $F = 0$, a void region is defined. This void region represents areas that are filled with a gas with a much lower density than the melted substrate. The gas dynamics in the void regions are not explicitly modeled, they are assigned a uniform pressure to reduce the computational expense of the simulation.

\begin{equation}\label{VOF}
    \frac{\partial F}{\partial t} + \left ( \nabla F \cdot v \right )= 0
\end{equation}

The laser beam is expressed as a Gaussian heat source, with a distribution described by
\begin{equation}
q = \frac{P}{r_0^2 \pi}\exp \left \{ - \left ( \frac{\sqrt{2} r}{r_0} \right )^2 \right \}
\end{equation}
where $P$ is the laser power and $r_0$ is the laser beam radius. During simulation, the heat flux is only applied to the top layer of surface cells and propagates through the rest of the domain through heat transfer mechanisms.

The evaporation mechanisms are also simulated, to model the ejection of vapor during the melting process. The recoil pressure is given by 
\begin{equation}
    P_s = a P_{V} \exp \left \{  \frac{\Delta H_v}{(\gamma - 1)C_v^{vap} T_v} \left ( 1 - \frac{T_v}{T}\right ) \right \}
\end{equation}

Here, the accommodation coefficient $a$, defines the ratio of the energy exchanged with the vapor phase, to the maximum energy exchange that is thermodynamically possible. $\Delta H_v$ represents the latent heat of vaporization, $\gamma$ is the ratio of specific heats, $C_v^{vap}$ is the constant volume heat capacity of the vapor, and $P_V$,  $T_V$ describe the saturation temperature and pressure respectively. Accordingly, the total mass loss, $\dot{M}$, due to this evaporation process is also taken into account, with the following relationship.
\begin{equation}
    \dot{M} = a \sqrt{\frac{1}{2 \pi R \bar{T}}}(P_l^{sat} - P_p)
\end{equation}
In the above equation, $R$ refers to the gas constant, $\bar{T}$ describes the average temperature of the liquid along the free surface, and $P_p$ is the partial pressure of the metal vapor. 

During the melting process, the keyhole that is formed in high energy density situations can increase the absorptivity of the melt pool, due to the laser beam reflecting within the keyhole cavity \cite{trapp2017situ}. To account for this effect, the relationship between the power, velocity, and absorptivity is determined based on a scaling relationship introduced in \cite{ye2019energy}. 

This scaling relationship reproduced in Equations \ref{absorptivity} and \ref{scaling} determines the absorptivity $A$ as a function of the beam radius, $r_0$, thermal diffusivity, $D$, melting enthalpy $H_m$, the minimum absorptivity on a flat plate, $A_m$, and the processing parameters. Generally, as the energy density of the laser increases, the keyhole deepens, and the local absorptivity increases.

\begin{equation}
    \label{absorptivity}
    A = \left (  0.70 (1 - e^{-0.66y}) \right )
\end{equation}

\begin{equation}
\label{scaling}
       y = \left ( \frac{A_m PD}{\pi v H_m r_0^2 \sqrt{Dr_0^2/v}} (r_0 \sqrt{D r_0/v})\right)
\end{equation}

This simulation was performed on a regular cartesian mesh, with mesh elements of size 5 $\mu m$. A complete list of the values for material parameters used to simulate the melt pool behavior can be found in the Appendix, in Tables \ref{table:materialparam} and \ref{table:materialparam2}.


Flow-3D uses a Volume-Of-Fluid (VOF) approach towards simulating the heat transfer dynamics, which depending on the resolution of the simulation, can cause isolated mesh elements to take on artificially high temperatures. This can result when the laser impacts small, disconnected ejected particles, or when the dynamics of the fluid flow are chaotic and simulated at relatively coarse resolutions. To resolve this issue, we take the 99.9\% percentile of the temperature values above the melting point and clipped the values above that temperature. The room temperature is 293K and the 99.9\% percentile was 6500K, so we use these values as $T_{min}$ and $T_{max}$ respectively. This filtering process allows us to use a mesh element size that is sufficiently coarse to enable the generation of large amounts of data while preserving the correct temperature distribution. A comparison between the generated Flow-3D data and reported experimental data from literature is made in Figure \ref{fig:result5}. This comparison is made for the melt pool depth for SS316L and the keyhole depth in Ti-6Al-4V. Both comparisons are made within the keyhole regime where analytical models such as the Rosenthal equation are no longer applicable \cite{promoppatum2017comprehensive}. This demonstrates that the scaled absorptivity is able to simulate the increase in melt pool depth caused by the multiple reflection effect in the keyhole. 

In order to feed the data to our network, we normalize the temperature values of the thermal field to be on the order $\sim 10^0$ for efficient training of the neural network model. Therefore, we apply a linear transformation to the temperature values to map $T_{min}$ and $T_{max}$ to 0 and 1 respectively, and rescale the other temperature values appropriately. We also crop the temperature field to a region of interest to train the model directly on the area where the melt pool is located at a given point in time. This modification reduces the amount of memory required for training and improves the learning process. 


\section{Methodology}
We begin this section by explaining the architecture of the main deep learning model of our method. Then, we continue by introducing an auxiliary network that improves the performance of the original model and how it is trained.
 
\subsection{Network architecture}
\begin{figure*}[htpb]    
\begin{center}
\includegraphics[width=1\linewidth]{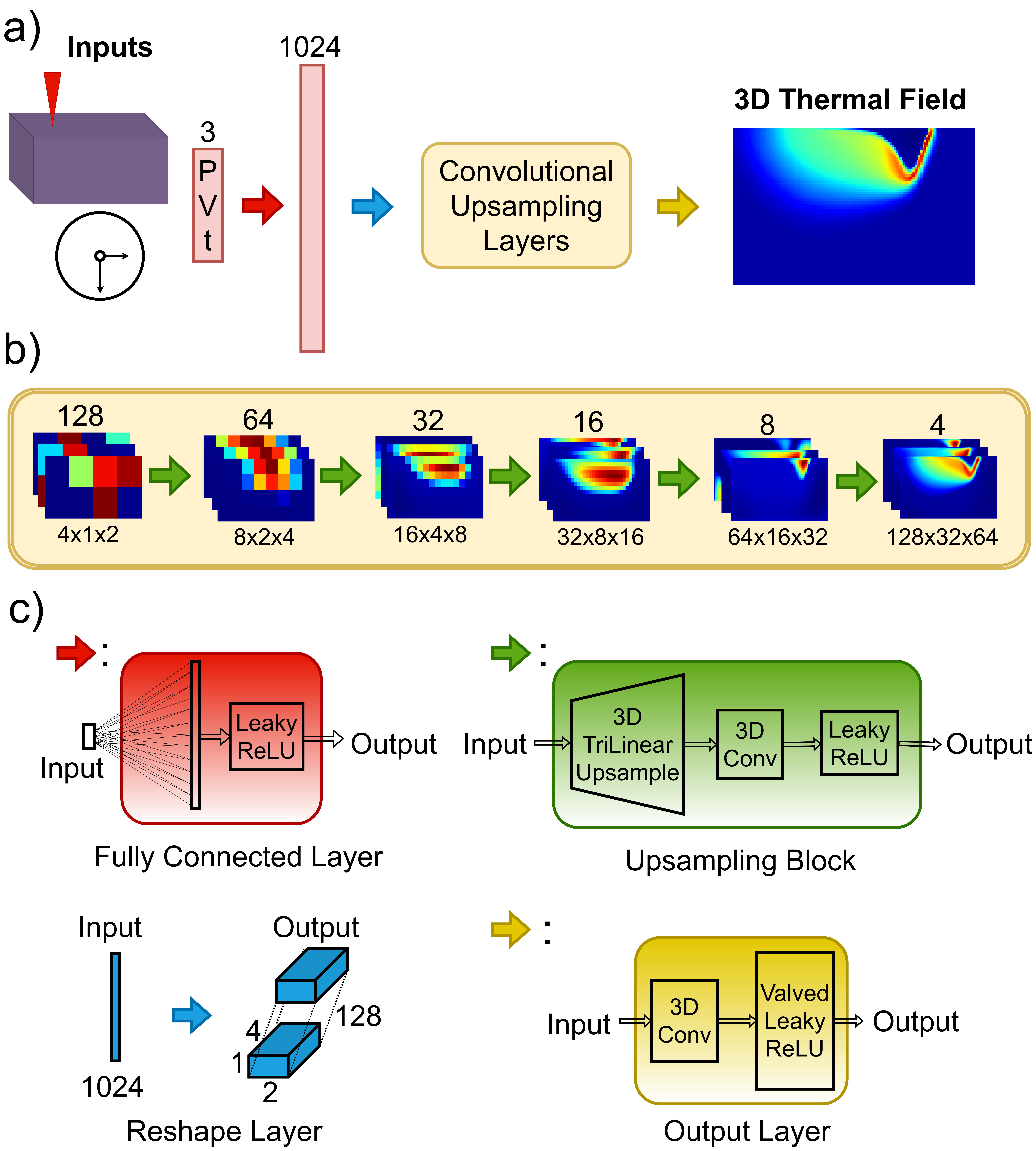}
\end{center}
\caption{a) The overall schematic of the architecture of T-CNN. b) The convolutional upsampling layers and the feature maps through the network. The number of channels and the dimensions of each feature map are noted above and below it respectively. c) The operators used in the network.}
\label{fig:upsample}
\end{figure*}

Figure \ref{fig:upsample}a provides a schematic of our model's architecture. Our Temperature CNN (T-CNN) consists of a fully connected layer followed by several 3D convolutional layers. The input vector simply contains the laser beam's power(P) and velocity(V) and time (t). After a  feature vector is constructed from the input by the fully connected layer, it is reshaped into a coarse-grid feature map. This can be interpreted as a learned feature vector for each point of a 3D coarse-grid tensor. This feature map is then passed through several upsampling and convolutional layers to increase the spatial resolution and decrease the number of channels. An illustration of the feature map's journey throughout the CNN can be found in Figure \ref{fig:upsample}b. The shapes of the feature maps are specific to the Ti64-5m dataset, but other aspects of the model remain the same for all datasets. All convolutional layers use kernel size 3 and stride 1 with 'same' padding. The scale factor for the trilinear upsampling layers is 2. We can observe that after the first two layers, the overall pattern and geometry of the melt pool are captured and some finer features are then learned in the following layers. We can also notice the keyhole's geometry as one of the learned feature maps in the second to last layer for this particular example.

Regarding the hyperparameters of our model, the first one is the number of channels of the first spatial feature map (128 in Figure \ref{fig:upsample}b). Choosing a lower number of channels (64, 32, ...) resulted in a noticeable deterioration in the performance of the model and bigger values (256, 512, ...) did not make a significant improvement, so we chose 128 as the final choice. There is also the choice of using a deeper fully connected network for the initial feature vector construction, but adding more hidden layers to the initial network did not lead to any significant improvement either. We also considered the option to use transposed convolution instead of non-parametric upsampling like the trilinear method, but it led to the checkerboard effect and no significant difference in the performance, so non-parametric upsampling was chosen to have a simpler and faster network. In order to find the best activation function, we tried using ELU, ReLU, SeLU, CELU, and Leaky ReLU, and all of them had a similar outcome, so we chose Leaky ReLU for faster optimization and lower computational cost. The final layer's activation function is called a "Valved" Leaky ReLU because it is a Leaky ReLU in the training phase but a ReLU in the testing phase. Therefore, we call it a valved Leaky ReLU since we open and close the leak manually. This is because negative output values are not physically possible (temperature lower than the ambient temperature), so we prevent the model from predicting negative outputs. However, we need to design the model in a way that it can learn to predict positive outputs, so we let the last activation be leaky in the training phase so that there is a gradient that can make the network learn not to predict negative values and tweak its parameters in the right way.

This problem is a regression task on the 3D thermal field, so we use the mean squared error (MSE) as the loss function to train our model. This loss is calculated by taking the average of the squared error across all mesh elements in the 3D thermal field. Therefore, the loss for a single sample is defined as follows:
\begin{equation}
    MSE = \frac{1}{LWD}\sum_{x=1}^{L}\sum_{y=1}^{W}\sum_{z=1}^{D}[\hat{T}(x,y,z)-T(x,y,z)]^2
\end{equation}
where $\hat{T}$ is the predicted temperature value at the mesh element with the positional index $[x,y,z]$ of the thermal field made by our T-CNN, $T$ is the value for the same point from the output of Flow-3D, and L, W, D represent the length, width, and depth of the cartesian mesh respectively. We use the Adam optimizer\cite{kingma2014adam} with a learning rate of 0.0002 to optimize the parameters of the network and minimize the loss function. We also utilize PyTorch's learning rate scheduler to decrease the learning rate when the learning curve reaches a plateau. The factor for the scheduler is set to 0.2 and patience is set to 3, meaning if the average training loss does not improve after 3 consecutive epochs, the learning rate is multiplied by 0.2.

\subsection{Using an auxiliary masker network}
After training T-CNN, we observed appealing results in terms of visual quality and achieved an average relative Root Mean Square Error of 2.5\% on predicting the thermal field. However, there is a delicate issue with our model which is caused by how the data is defined. If you pay close attention to the output of the model in Figure \ref{fig:upsample}a, you may notice a thin layer in the boundary of the keyhole, which is visible between the empty area of the keyhole and the high temperature values of the melt pool. The reason behind this is the value of the thermal field in the keyhole space. The temperature value that Flow-3D assigns to the points not occupied by metal particles is equal to the room temperature. Since the output of T-CNN is continuous, it cannot suddenly change from a few thousand Kelvins to room temperature. Therefore, we see a thin layer with intermediate temperature values between room temperature and very hot values since the temperature decreases gradually over a few mesh elements.
\begin{figure*}[htpb]    
\begin{center}
\includegraphics[width=1\linewidth]{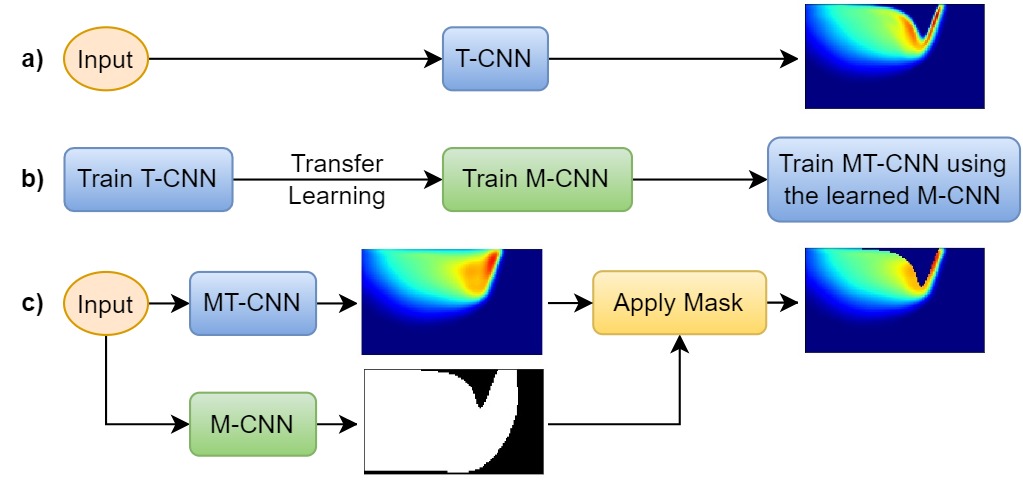}
\end{center}
\caption{A simple illustration of the role of the auxiliary M-CNN. a) The initial T-CNN model's functionality. b) How M-CNN and TM-CNN are trained. c) The MT-CNN model augmented with M-CNN}
\label{fig:mask}
\end{figure*}

In order to make the output of the model more realistic and avoid creating this thin layer, we use our pre-trained T-CNN and train an auxiliary network called the masker CNN (M-CNN). The architecture of this model is the same as Figure \ref{fig:upsample}a, except that the final layer's activation is the sigmoid function. The job of M-CNN is to predict a binary mask for the keyhole. To be more precise, its output is supposed to estimate the areas which are assigned the exact value of room temperature by Flow-3D. That includes both the keyhole space and the regions of the metal where the heat has not reached yet. After training the M-CNN to learn this task, we train a new Masked Temperature CNN (MT-CNN) to predict the thermal field. Now, the values in the keyhole space are set to room temperature by the mask provided by M-CNN, and MT-CNN's output is neglected in that space. Therefore, MT-CNN can freely predict high temperature values in that region without having to try to predict the room temperature for the keyhole at the same time. A simple comparison between T-CNN and MT-CNN and their outputs can be found in Figure \ref{fig:mask}.

\section{Results and discussion}

\begin{figure*}[htpb]    
\begin{center}
\includegraphics[width=1\linewidth]{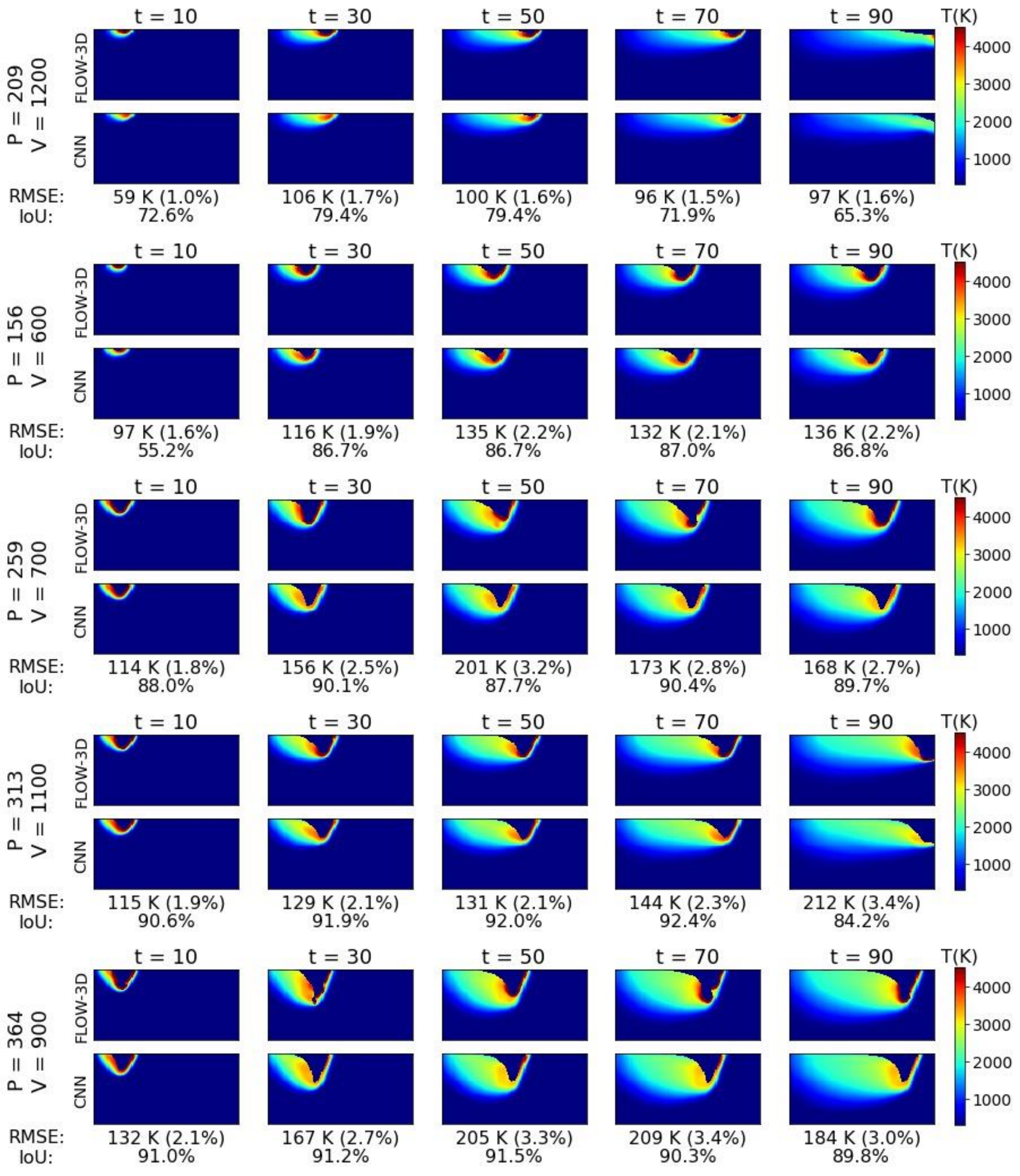}
\end{center}
\caption{Performance of MT-CNN on some validation samples from the Ti64-5m dataset. Note that the original data is 3D, and the cross-section of the thermal field is shown in these plots.}
\label{fig:samples1}
\end{figure*}

We begin with a qualitative comparison between our model's predictions and the results of the direct numerical simulations obtained from Flow-3D. Figure \ref{fig:samples1} contains a few of the validation samples from the Ti64-5m dataset. We can see that the predicted thermal field closely matches the ground truth for the most part. Quantitative metrics will be explained later. We can say that the only location where the model's prediction noticeably differs from the ground truth is the keyhole area. This is because this area contains the most turbulent and transient behavior compared to the other parts of the melt pool, especially in high energy density regimes. Since the model is trained with a simple MSE loss, it only learns the dominant pattern and geometry of the thermal field and fails to capture the turbulent phenomena happening at the keyhole. However, learning the overall geometry is indeed the main purpose of this work, because important features like the keyhole's depth can be obtained from it. After all, losing unimportant information is the trade-off that we are willing to take to obtain the important information more quickly and efficiently.

\renewcommand{\arraystretch}{1.5}
\begin{table}[htbp]
\caption{The mean and standard deviation of the quantitative evaluation metrics.}
\begin{center}
\begin{tabular}{|c|c|c|c|c|}
\hline
& \multicolumn{2}{|c|}{\textbf{T-CNN}} & \multicolumn{2}{|c|}{\textbf{MT-CNN}}\\
\hline
\textbf{Dataset}  & \textbf{RMSE} & \textbf{IoU} & \textbf{RMSE} & \textbf{IoU}\\
\hline
Ti64-5m & $2.55\pm1.04 \%$ & $81.59\pm14.67 \%$ & $2.81\pm1.21 \%$ & $84.34\pm12.46 \%$ \\
\hline
Ti64-10m & $2.18\pm1.25 \%$ & $79.17\pm15.52 \%$ & $2.45\pm1.39 \%$ & $80.53\pm14.09 \%$ \\
\hline
Ti64-10m-p & $2.00\pm0.79 \%$ & $83.33\pm10.06 \%$ & $2.19\pm0.89 \%$ & $85.26\pm7.82 \%$ \\
\hline
SS316L-5m-p & $3.37\pm1.11 \%$  & $81.48\pm10.51 \%$ & $3.73\pm1.30 \%$ & $81.86\pm8.31 \%$ \\
\hline
\end{tabular}
\label{table:result}
\end{center}
\end{table}

For the quantitative evaluation of our model's performance, we use two metrics: Root-Mean-Square Error (RMSE) on the thermal field and Intersection over Union (IoU) on the melt pool. We consider each mesh element with a temperature value higher than the melting point as liquid. The melting point is set as the average of liquidus and solidus temperature, which are provided in Tables \ref{table:materialparam} and \ref{table:materialparam2}. Looking at Figure \ref{fig:samples1}, we can see that the metrics achieved by the model also indicate its success. The performance of the model for other datasets is summarized in Table \ref{table:result}. The performance of T-CNN is also included for the sake of comparison. We can see that despite the increase in the RMSE, the IoU score is improved by MT-CNN.

\begin{figure*}[htpb]
\begin{center}
\includegraphics[width=1\linewidth]{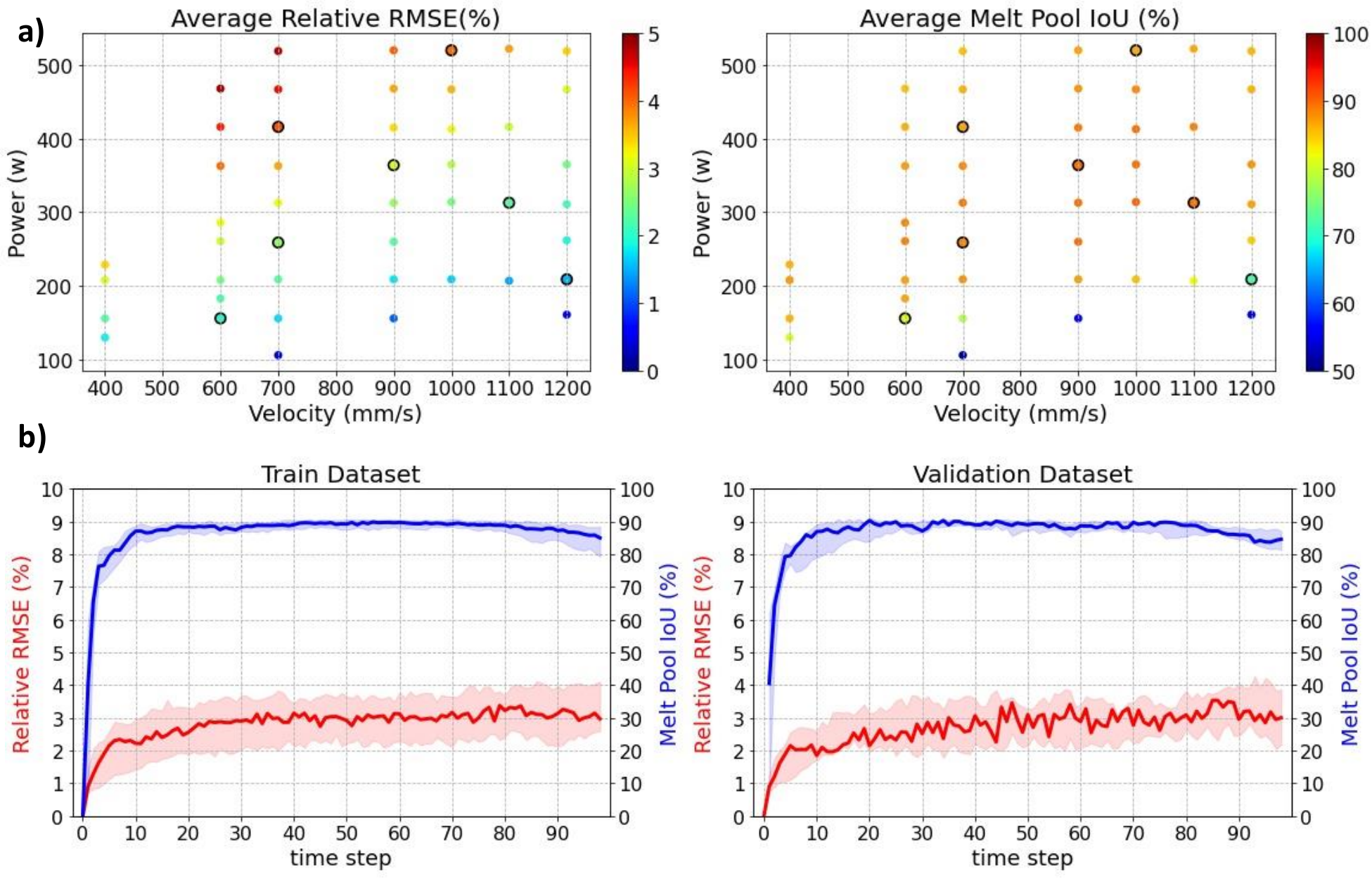}
\end{center}
\caption{The summarized results for Ti64-5m dataset. a) The average performance of the model over the data distribution. The test samples are marked by a black edge.  b) The performance of the model over time. The main line indicates the median value for the metric, and the shaded area indicates the interquartile range.}
\label{fig:result1}
\end{figure*}

In order to investigate the performance of the model further, we will look into the distribution of our evaluation metrics over time and the data distribution. Figure \ref{fig:result1}a shows the averaged value of the metrics over time for each sample of the dataset. We can see that there is a noticeable correlation between the performance of the model and the energy density. In the bottom right of these plots, the energy density is low and the melt pool is relatively small. Since most of the thermal field is still unaffected by the heat from the laser beam, both the ground truth and the prediction are still mostly at room temperature, therefore the RMSE is lower in this regime. It is also observed that the IoU score is lower than the other regions of the plot which is due to the small size of the melt pool. With a small melt pool, the IoU score is affected more severely by false predictions of the model. Moving to the high energy density regime in the upper left of the plots, we can observe an increase in the RMSE and the IoU score because of the larger size of the melt pool. The turbulence in the keyhole greatly increases the RMSE in these regimes, but the IoU score is still high and the model has succeeded in estimating the melt pool's overall geometry. 

Figure \ref{fig:result1}b shows the distribution of the metrics over the dataset for each time step. We can see that both the RMSE and the IoU score are much lower in the initial time steps. This is because of the small size of the melt pool as explained before. The IoU score is very low at the beginning of the process which decreases the overall average for the metric. However, we can see that the model achieves an IoU score of 90\% for most time steps which again shows its successful estimation of the melt pool geometry. The slight decrease of IoU in the last few time steps is due to the solidification of the metal which happens only for a few samples where the velocity is too high and the laser beam exits the modeled part. Since there is very little data on this phenomenon, the model has not learned to predict the thermal field in such a situation compared to the earlier time steps.

In this section, we compared the predictions of our model and the ground truth for a qualitative evaluation of MT-CNN and further analyzed the results by looking closer into the distribution of the RMSE and the IoU score over the dataset and through time. We chose the Ti64-5m dataset to elaborate on in this section because the model successfully achieved its goal despite having only 46 sampled processes and the highest output dimensionality among other datasets. Similar plots and figures for the other datasets can be found in \ref{appendix:results}. The model achieves similar results for all datasets, but it does not work as well for the SS-5m-5 dataset. This can be because of the low number of data samples and also the higher amount of turbulent phenomena in this dataset. However, we can see that our model still succeeds in learning the main geometry of the melt pool in all datasets by looking at the IoU score and comparing the predictions with the ground truth.

\section{Conclusion and future work}
In this paper, we proposed a deep learning framework that successfully predicts the thermal field and the dominant melt pool geometry in LPBF additive manufacturing processes. Compared to the Flow-3D software which uses iterative numerical solvers, this model has a huge computational advantage. Since time is included as one of the inputs to our CNN, we can obtain the thermal field at any arbitrary time instantly without the need to run a whole simulation. For the Ti-6Al-4V alloy, the model achieved an average relative RMSE of 2 to 3\% on predicting the thermal field and an average IoU of 80 to 90\% on predicting the melt pool area. We had a limited dataset for the SS316L dataset and the model achieved a relative RMSE of 3.7\% and an IoU score of 82\% which is not as good as Ti-6Al-4V. However, the model successfully captures the dominant pattern and features of the melt pool for all datasets used in this work.
\bibliographystyle{elsarticle-num} 
\bibliography{cas-refs}

\begin{thebibliography}{10}
\expandafter\ifx\csname url\endcsname\relax
  \def\url#1{\texttt{#1}}\fi
\expandafter\ifx\csname urlprefix\endcsname\relax\def\urlprefix{URL }\fi
\expandafter\ifx\csname href\endcsname\relax
  \def\href#1#2{#2} \def\path#1{#1}\fi

\bibitem{debroy2018additive}
T.~DebRoy, H.~Wei, J.~Zuback, T.~Mukherjee, J.~Elmer, J.~Milewski, A.~M. Beese,
  A.~d. Wilson-Heid, A.~De, W.~Zhang, Additive manufacturing of metallic
  components--process, structure and properties, Progress in Materials Science
  92 (2018) 112--224.

\bibitem{jiang2020path}
J.~Jiang, Y.~Ma, Path planning strategies to optimize accuracy, quality, build
  time and material use in additive manufacturing: a review, Micromachines
  11~(7) (2020) 633.

\bibitem{AKBARI2022102817}
P.~Akbari, F.~Ogoke, N.-Y. Kao, K.~Meidani, C.-Y. Yeh, W.~Lee, A.~B. Farimani,
  \href{https://www.sciencedirect.com/science/article/pii/S2214860422002172}{Meltpoolnet:
  Melt pool characteristic prediction in metal additive manufacturing using
  machine learning}, Additive Manufacturing (2022) 102817\href
  {https://doi.org/https://doi.org/10.1016/j.addma.2022.102817}
  {\path{doi:https://doi.org/10.1016/j.addma.2022.102817}}.
\newline\urlprefix\url{https://www.sciencedirect.com/science/article/pii/S2214860422002172}

\bibitem{WANG2020101538}
C.~Wang, X.~Tan, S.~Tor, C.~Lim, Machine learning in additive manufacturing:
  State-of-the-art and perspectives, Additive Manufacturing 36 (2020) 101538.
\newblock \href {https://doi.org/https://doi.org/10.1016/j.addma.2020.101538}
  {\path{doi:https://doi.org/10.1016/j.addma.2020.101538}}.

\bibitem{FLOW-3D}
I.~Flow~Science, \href{https://www.flow3d.com/}{FLOW-3D, Version~12.0}, Santa
  Fe, NM (2019).
\newline\urlprefix\url{https://www.flow3d.com/}

\bibitem{yan2018data}
F.~Yan, Y.-C. Chan, A.~Saboo, J.~Shah, G.~B. Olson, W.~Chen, Data-driven
  prediction of mechanical properties in support of rapid certification of
  additively manufactured alloys, Computer Modeling in Engineering \& Sciences
  117~(3) (2018) 343--366.

\bibitem{krizhevsky2012imagenet}
A.~Krizhevsky, I.~Sutskever, G.~E. Hinton, Imagenet classification with deep
  convolutional neural networks, Advances in neural information processing
  systems 25 (2012).

\bibitem{mnih2015human}
V.~Mnih, K.~Kavukcuoglu, D.~Silver, A.~A. Rusu, J.~Veness, M.~G. Bellemare,
  A.~Graves, M.~Riedmiller, A.~K. Fidjeland, G.~Ostrovski, et~al., Human-level
  control through deep reinforcement learning, nature 518~(7540) (2015)
  529--533.

\bibitem{voulodimos2018deep}
A.~Voulodimos, N.~Doulamis, A.~Doulamis, E.~Protopapadakis, Deep learning for
  computer vision: A brief review, Computational intelligence and neuroscience
  2018 (2018).

\bibitem{hinton2012deep}
G.~Hinton, L.~Deng, D.~Yu, G.~E. Dahl, A.-r. Mohamed, N.~Jaitly, A.~Senior,
  V.~Vanhoucke, P.~Nguyen, T.~N. Sainath, et~al., Deep neural networks for
  acoustic modeling in speech recognition: The shared views of four research
  groups, IEEE Signal processing magazine 29~(6) (2012) 82--97.

\bibitem{deng2013new}
L.~Deng, G.~Hinton, B.~Kingsbury, New types of deep neural network learning for
  speech recognition and related applications: An overview, in: 2013 IEEE
  international conference on acoustics, speech and signal processing, IEEE,
  2013, pp. 8599--8603.

\bibitem{graves2013speech}
A.~Graves, A.-r. Mohamed, G.~Hinton, Speech recognition with deep recurrent
  neural networks, in: 2013 IEEE international conference on acoustics, speech
  and signal processing, Ieee, 2013, pp. 6645--6649.

\bibitem{abdel2014convolutional}
O.~Abdel-Hamid, A.-r. Mohamed, H.~Jiang, L.~Deng, G.~Penn, D.~Yu, Convolutional
  neural networks for speech recognition, IEEE/ACM Transactions on audio,
  speech, and language processing 22~(10) (2014) 1533--1545.

\bibitem{lecun1995convolutional}
Y.~LeCun, Y.~Bengio, et~al., Convolutional networks for images, speech, and
  time series, The handbook of brain theory and neural networks 3361~(10)
  (1995) 1995.

\bibitem{simonyan2014very}
K.~Simonyan, A.~Zisserman, Very deep convolutional networks for large-scale
  image recognition, arXiv preprint arXiv:1409.1556 (2014).

\bibitem{long2015fully}
J.~Long, E.~Shelhamer, T.~Darrell, Fully convolutional networks for semantic
  segmentation, in: Proceedings of the IEEE conference on computer vision and
  pattern recognition, 2015, pp. 3431--3440.

\bibitem{karpathy2014large}
A.~Karpathy, G.~Toderici, S.~Shetty, T.~Leung, R.~Sukthankar, L.~Fei-Fei,
  Large-scale video classification with convolutional neural networks, in:
  Proceedings of the IEEE conference on Computer Vision and Pattern
  Recognition, 2014, pp. 1725--1732.

\bibitem{kutz2017deep}
J.~N. Kutz, Deep learning in fluid dynamics, Journal of Fluid Mechanics 814
  (2017) 1--4.

\bibitem{kirchdoerfer2016data}
T.~Kirchdoerfer, M.~Ortiz, Data-driven computational mechanics, Computer
  Methods in Applied Mechanics and Engineering 304 (2016) 81--101.

\bibitem{oishi2017computational}
A.~Oishi, G.~Yagawa, Computational mechanics enhanced by deep learning,
  Computer Methods in Applied Mechanics and Engineering 327 (2017) 327--351.

\bibitem{sirignano2018dgm}
J.~Sirignano, K.~Spiliopoulos, Dgm: A deep learning algorithm for solving
  partial differential equations, Journal of computational physics 375 (2018)
  1339--1364.

\bibitem{han2018solving}
J.~Han, A.~Jentzen, E.~Weinan, Solving high-dimensional partial differential
  equations using deep learning, Proceedings of the National Academy of
  Sciences 115~(34) (2018) 8505--8510.

\bibitem{raissi2019physics}
M.~Raissi, P.~Perdikaris, G.~E. Karniadakis, Physics-informed neural networks:
  A deep learning framework for solving forward and inverse problems involving
  nonlinear partial differential equations, Journal of Computational physics
  378 (2019) 686--707.

\bibitem{liang2018deep}
L.~Liang, M.~Liu, C.~Martin, W.~Sun, A deep learning approach to estimate
  stress distribution: a fast and accurate surrogate of finite-element
  analysis, Journal of The Royal Society Interface 15~(138) (2018) 20170844.

\bibitem{jin2020deep}
Z.~L. Jin, Y.~Liu, L.~J. Durlofsky, Deep-learning-based surrogate model for
  reservoir simulation with time-varying well controls, Journal of Petroleum
  Science and Engineering 192 (2020) 107273.

\bibitem{tang2020deep}
M.~Tang, Y.~Liu, L.~J. Durlofsky, A deep-learning-based surrogate model for
  data assimilation in dynamic subsurface flow problems, Journal of
  Computational Physics 413 (2020) 109456.

\bibitem{li2020reaction}
A.~Li, R.~Chen, A.~B. Farimani, Y.~J. Zhang, Reaction diffusion system
  prediction based on convolutional neural network, Scientific reports 10~(1)
  (2020) 1--9.

\bibitem{jiang2020deep}
C.~Jiang, A.~B. Farimani, Deep learning convective flow using conditional
  generative adversarial networks, arXiv preprint arXiv:2005.06422 (2020).

\bibitem{mills2002recommended}
K.~C. Mills, Recommended values of thermophysical properties for selected
  commercial alloys, Woodhead Publishing, 2002.

\bibitem{hirt1981volume}
C.~W. Hirt, B.~D. Nichols, Volume of fluid (vof) method for the dynamics of
  free boundaries, Journal of computational physics 39~(1) (1981) 201--225.

\bibitem{trapp2017situ}
J.~Trapp, A.~M. Rubenchik, G.~Guss, M.~J. Matthews, In situ absorptivity
  measurements of metallic powders during laser powder-bed fusion additive
  manufacturing, Applied Materials Today 9 (2017) 341--349.

\bibitem{ye2019energy}
J.~Ye, S.~A. Khairallah, A.~M. Rubenchik, M.~F. Crumb, G.~Guss, J.~Belak, M.~J.
  Matthews, Energy coupling mechanisms and scaling behavior associated with
  laser powder bed fusion additive manufacturing, Advanced Engineering
  Materials 21~(7) (2019) 1900185.

\bibitem{promoppatum2017comprehensive}
P.~Promoppatum, S.-C. Yao, P.~C. Pistorius, A.~D. Rollett, A comprehensive
  comparison of the analytical and numerical prediction of the thermal history
  and solidification microstructure of inconel 718 products made by laser
  powder-bed fusion, Engineering 3~(5) (2017) 685--694.

\bibitem{kingma2014adam}
D.~P. Kingma, J.~Ba, Adam: A method for stochastic optimization, arXiv preprint
  arXiv:1412.6980 (2014).

\bibitem{forien2020detecting}
J.-B. Forien, N.~P. Calta, P.~J. DePond, G.~M. Guss, T.~T. Roehling, M.~J.
  Matthews, Detecting keyhole pore defects and monitoring process signatures
  during laser powder bed fusion: A correlation between in situ pyrometry and
  ex situ x-ray radiography, Additive Manufacturing 35 (2020) 101336.

\bibitem{cunningham2019keyhole}
R.~Cunningham, C.~Zhao, N.~Parab, C.~Kantzos, J.~Pauza, K.~Fezzaa, T.~Sun,
  A.~D. Rollett, Keyhole threshold and morphology in laser melting revealed by
  ultrahigh-speed x-ray imaging, Science 363~(6429) (2019) 849--852.

\end{thebibliography}

\newpage
\appendix
\section{Material Parameters}
\begin{table}[ht!]
\caption{Material Parameters used to simulate the Ti-6Al-4V melting process}
\begin{tabular}{@{}lllll@{}}
\toprule
Parameter                        & Value                                            & Units                   &  &  \\ \midrule
Density, $\rho$, 298 K              & 4420                                             & kg/m$^3$ &  &  \\
Density, $\rho$, 1923 K             & 3920                                             & kg/m$^3$ &  &  \\
Specific Heat, $C_v$, 298 K         & 546                                              & J/kg/K                  &  &  \\
Specific Heat, $C_v$, 1923 K        & 831                                              & J/kg/K                  &  &  \\
Vapor Specific Heat, $C_{v}, vapor$ & 600                                              & J/kg/K                  &  &  \\
Thermal Conductivity, $k$, 298 K    & 7                                                & W/m/K                   &  &  \\
Thermal Conductivity, $k$, 1923 K  & 33.4                                             & W/m/K                   &  &  \\
Viscosity                        & 0.00325                                          & kg/m/s                  &  &  \\
Liquidus Temperature, $T_L$       & 1923                                             & K                       &  &  \\
Solidus Temperature, $T_S$        & 1873                                             & K                       &  &  \\
Latent heat of fusion            & 2.86 $\times 10^5$ & J/kg                    &  &  \\
Latent heat of vaporization     & 6.00 $\times 10^4$ & J/kg                    &  &  \\ \bottomrule
\end{tabular}
\label{table:materialparam}
\end{table}

\begin{table}[ht!]
\caption{Material Parameters used to simulate the SS316L melting process}
\begin{tabular}{@{}lllll@{}}
\toprule
Parameter                        & Value                                            & Units                   &  &  \\ \midrule
Density, $\rho$, 298 K              & 7950                                             & kg/m$^3$ &  &  \\
Density, $\rho$, 1923 K             & 7249                                             & kg/m$^3$ &  &  \\
Specific Heat, $C_v$, 298 K         & 470                                              & J/kg/K                  &  &  \\
Specific Heat, $C_v$, 1923 K        & 726                                              & J/kg/K                  &  &  \\
Vapor Specific Heat, $C_{v}, vapor$ & 600                                              & J/kg/K                  &  &  \\
Thermal Conductivity, $k$, 298 K    & 13.4                                                & W/m/K                   &  &  \\
Thermal Conductivity, $k$, 1923 K  & 29.0                                            & W/m/K                   &  &  \\
Viscosity                        & 0.008                                          & kg/m/s                  &  &  \\
Liquidus Temperature, $T_L$       & 1694                                             & K                       &  &  \\
Solidus Temperature, $T_S$        & 1717                                             & K                       &  &  \\
Latent heat of fusion            & 2.6 $\times 10^5$ & J/kg                    &  &  \\
Latent heat of vaporization     & 6.00 $\times 10^4$ & J/kg                    &  &  \\ \bottomrule
\end{tabular}
\label{table:materialparam2}
\end{table}

\clearpage
\section{Results for the rest of the datasets}
\label{appendix:results}
\begin{figure*}[!htpb]    
\begin{center}
\includegraphics[width=1\linewidth]{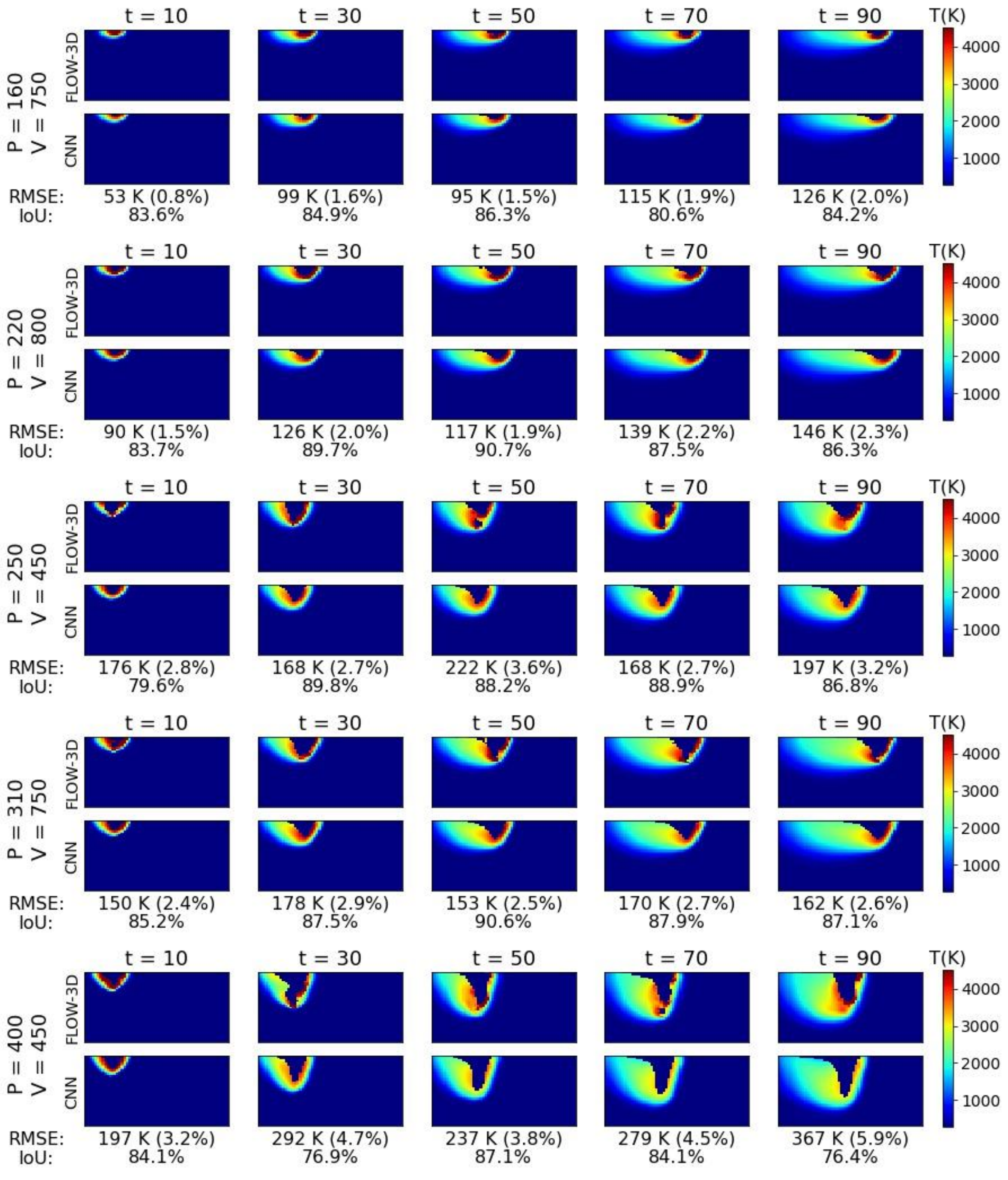}
\end{center}
\caption{Performance of MT-CNN on some validation samples from the Ti64-10m dataset.}
\label{fig:samples2}
\end{figure*}
\begin{figure*}[!htpb]    
\begin{center}
\includegraphics[width=1\linewidth]{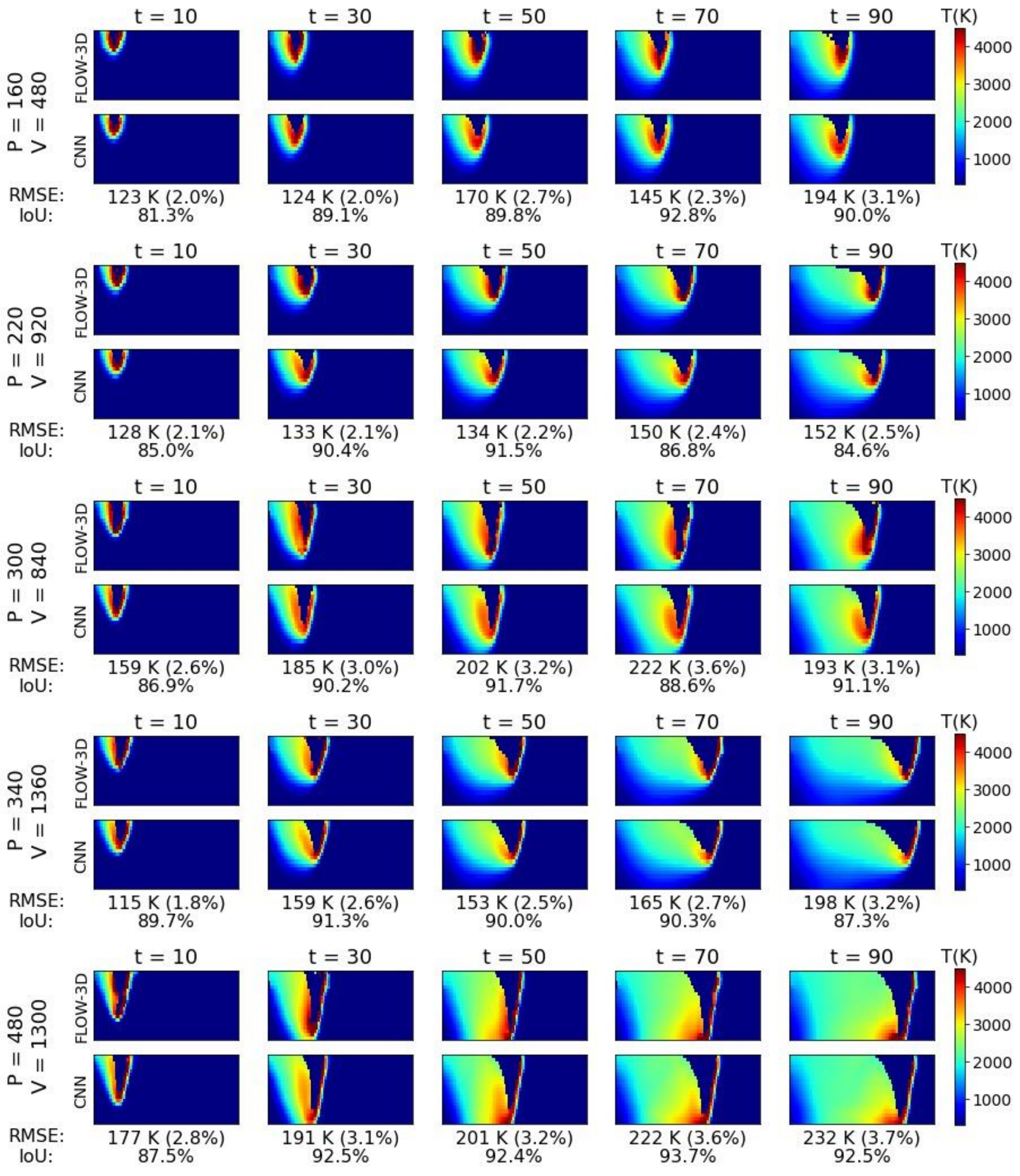}
\end{center}
\caption{Performance of MT-CNN on some validation samples from the Ti64-10m-p dataset.}
\label{fig:samples3}
\end{figure*}
\begin{figure*}[!htpb]    
\begin{center}
\includegraphics[width=1\linewidth]{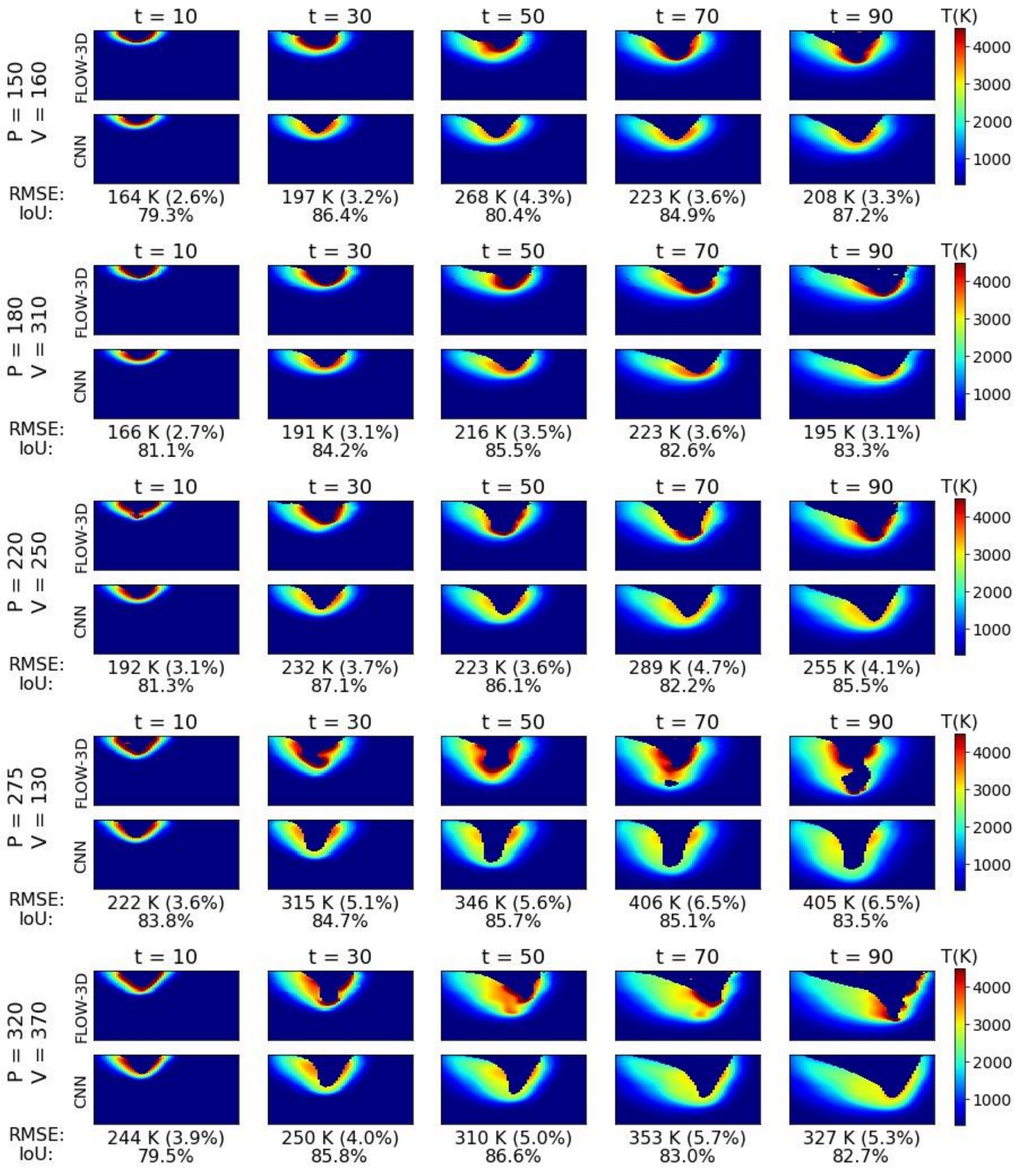}
\end{center}
\caption{Performance of MT-CNN on some validation samples from the SS-5m-p dataset.}
\label{fig:samples4}
\end{figure*}
\begin{figure*}[!htpb]    
\begin{center}
\includegraphics[width=1\linewidth]{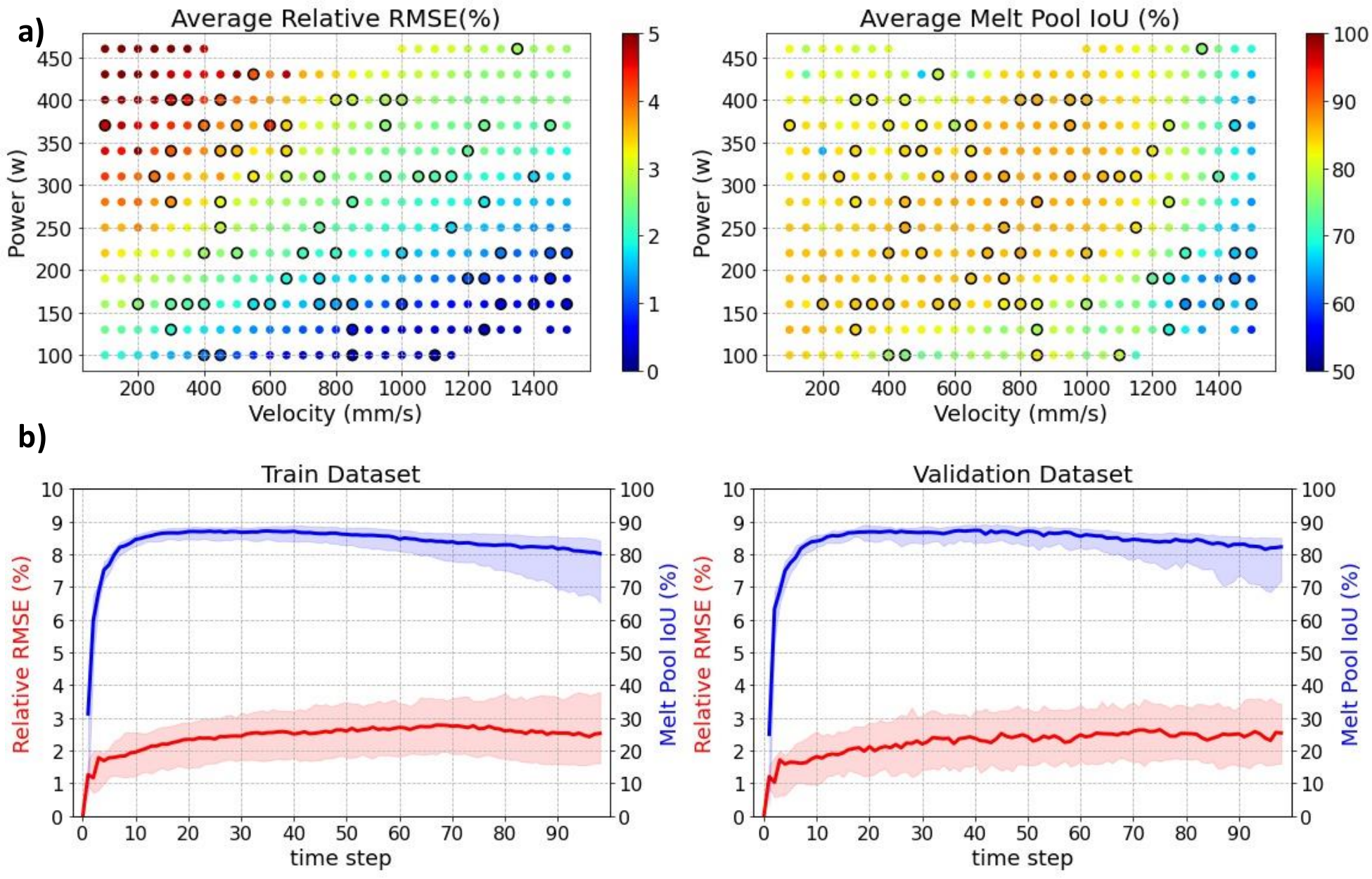}
\end{center}
\caption{Result for Ti64-10m dataset. a) Performance over the data distribution. The validation data is marked by a black edge. b) Performance over time.}
\label{fig:result2}
\end{figure*}
\begin{figure*}[!htpb]    
\begin{center}
\includegraphics[width=1\linewidth]{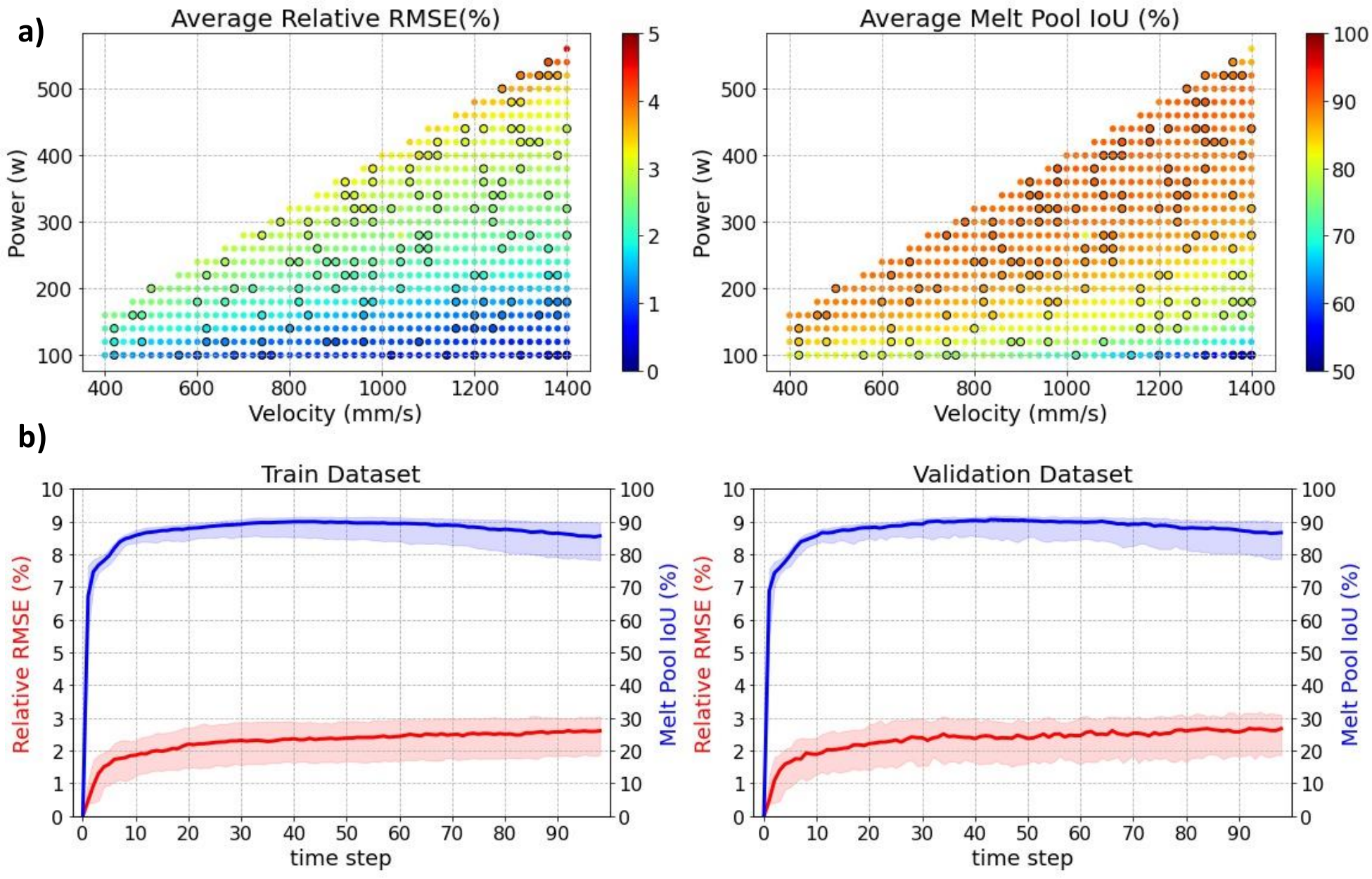}
\end{center}
\caption{Result for Ti64-10m-p dataset. a) Performance over the data distribution. The validation data is marked by a black edge. b) Performance over time.}
\label{fig:result3}
\end{figure*}
\begin{figure*}[!htpb]    
\begin{center}
\includegraphics[width=1\linewidth]{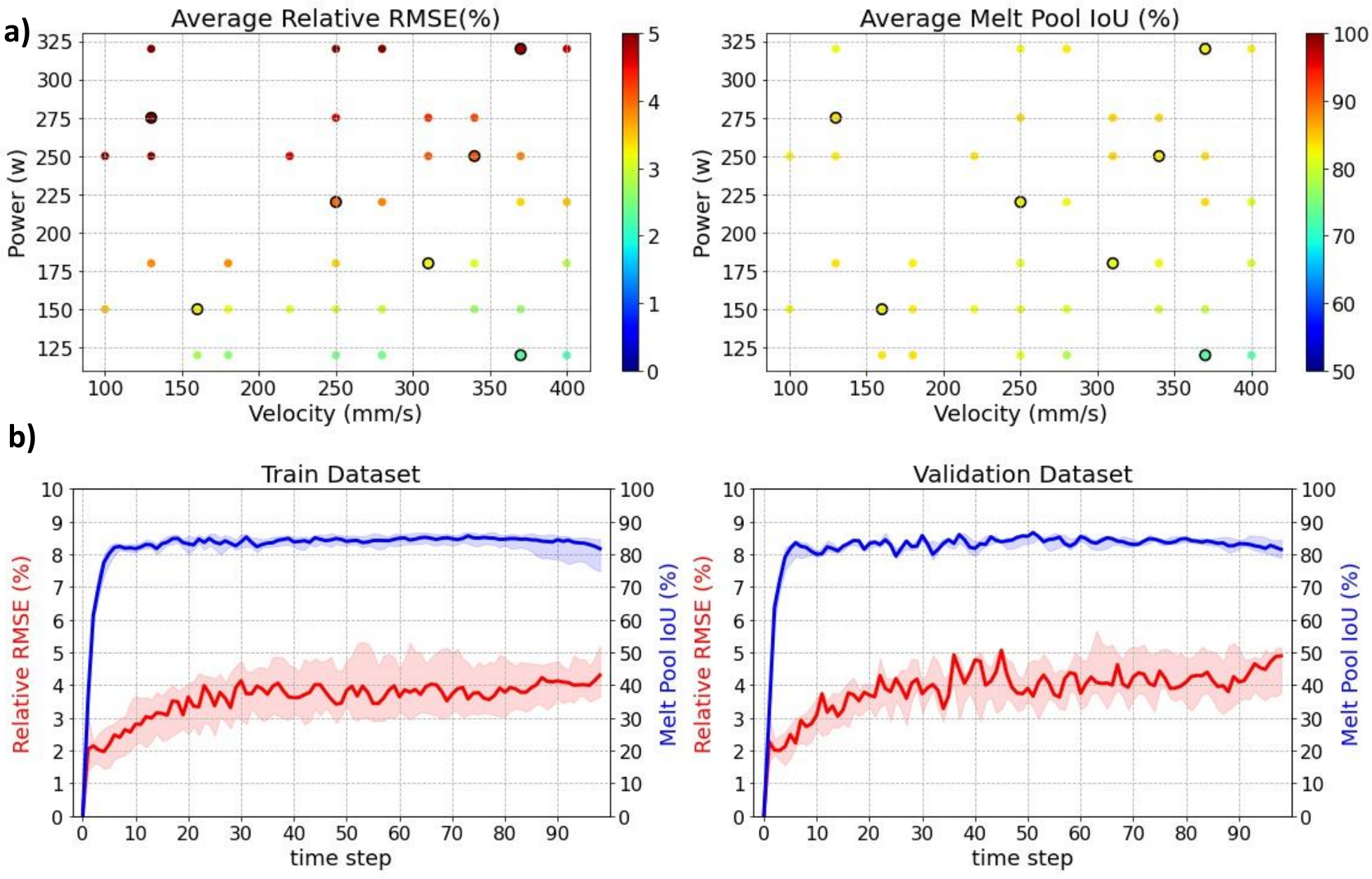}
\end{center}
\caption{Result for SS-5m-p dataset. a) Performance over the data distribution. The validation data is marked by a black edge. b) Performance over time.}
\label{fig:result4}
\end{figure*}

\clearpage
\section{Comparison of Flow-3D data with experimental data}
\begin{figure*}[!htpb]    
\begin{center}
\includegraphics[width=1\linewidth]{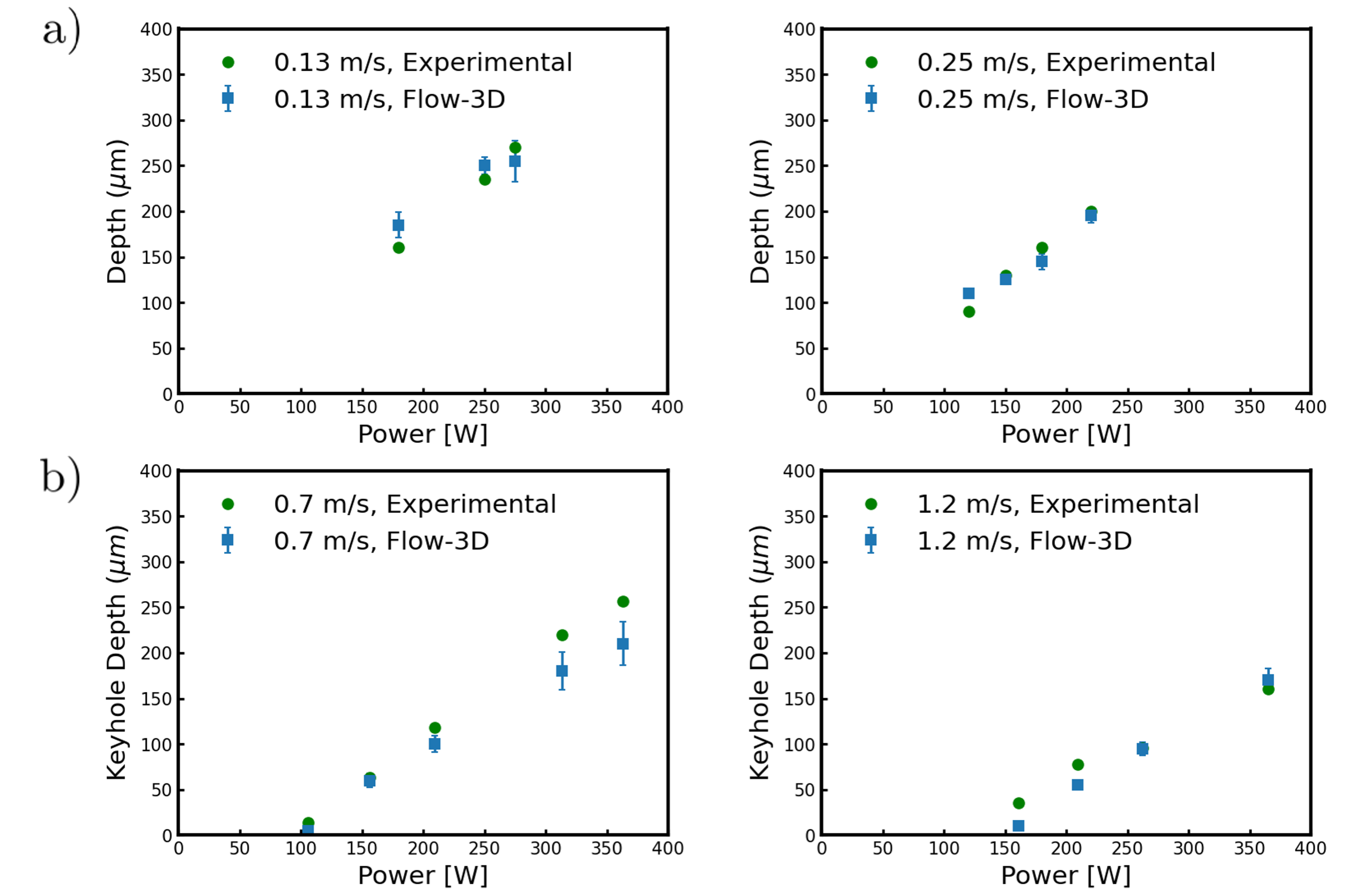}
\end{center}
\caption{A comparison of the Flow-3D simulated melt pool to experimental data from the literature. (a) Comparisons for the melt pool depth at varying powers at a 95 micron beam diameter, experimental data referenced from \cite{forien2020detecting} for SS316L. (b) Comparisons for the keyhole depth at varying powers at a 95 micron beam diameter, experimental data referenced from \cite{cunningham2019keyhole} for Ti-6Al-4V.}
\label{fig:result5}
\end{figure*}





\end{document}